\documentclass[]{spie}  

\usepackage{amsmath,amsfonts,amssymb}
\usepackage{graphicx}
\usepackage[colorlinks=true, allcolors=blue]{hyperref}

\title{Hardware based Spatio-Temporal Neural Processing Backend for Imaging Sensors: Towards a Smart Camera}

\author{Samiran Ganguly}
\author{Yunfei Gu}
\author{Mircea R. Stan}
\author{Avik W. Ghosh}
\affil{Charles L. Brown Dept. of Electrical and Computer Engineering\\University of Virginia, Charlottesville, VA 22904}

\authorinfo{Further author information: (Send correspondence to Samiran Ganguly)\\Samiran Ganguly: E-mail: sganguly@virginia.edu\\  Mircea R. Stan: E-mail: mircea@virginia.edu\\ Avik W. Ghosh: E-mail: ag7rq@virginia.edu}

\begin{document} 
\maketitle

\begin{abstract}
In this work we show how we can build a technology platform for cognitive imaging sensors using recent advances in recurrent neural network architectures and training methods inspired from biology. We demonstrate learning and processing tasks specific to imaging sensors, including enhancement of sensitivity and signal-to-noise ratio (SNR) purely through neural filtering beyond the fundamental limits sensor materials, and inferencing and spatio-temporal pattern recognition capabilities of these networks with applications in object detection, motion tracking and prediction. We then show designs of unit hardware cells built using complementary metal-oxide semiconductor (CMOS) and emerging materials technologies for ultra-compact and energy-efficient embedded neural processors for smart cameras.  
\end{abstract}


\section{INTRODUCTION}
\label{sec:intro}  

Last three decades have seen the progress of Neural Networks from a statistician’s playbook to the technology behemoth running the modern information and communications technology (ICT) industry. This has largely been made possible by advances in very large scale integration (VLSI) and transistor scaling providing evermore powerful hardware to run neural network software. This scaling is slowing down and heading towards real physical limits of atomic dimensions. As a result, the Moore's Law is now being reinterpreted as a call for deep multi-functional integration of erstwhile loosely coupled sub-systems: sensing, memory, and logic, in a single computing substrate with a resulting increase in the ``user value''. In this work, we illustrate how we can leverage these advances to build a technology platform for truly cognitive imaging sensors, inspired from biology. 

We first discuss recurrent neural network based architectures designed to process data with features laying in both spatial and temporal dimensions (e.g. video) and associated learning techniques, in particular: Convolutional Neural Network (CNN), Reservoir Computing (RC), and Hierarchical Temporal Memory (HTM), whose hardware implementation and operation are feasible with present day technology. We then demonstrate, using simulations, the use of these networks for learning and processing tasks specific to imaging sensors: a) Enhancing sensitivity and SNR/D* purely through neural filtering that goes beyond the fundamental limits of detection set by the material properties. b) Spatio-Temporal inferencing and pattern recognition capabilities built within these networks that can be used for object detection, motion tracking, and prediction. 

We then conclude by illustrating how unit hardware cells can be designed using both conventional CMOS as well as emerging nano-materials based technologies, such as spintronics and memristors, which can enable development of ultra-compact and energy-efficient embedded spatio-temporal neural processors for smart cameras.

 \begin{figure} [h]
 \begin{center}
 \includegraphics[width=5in]{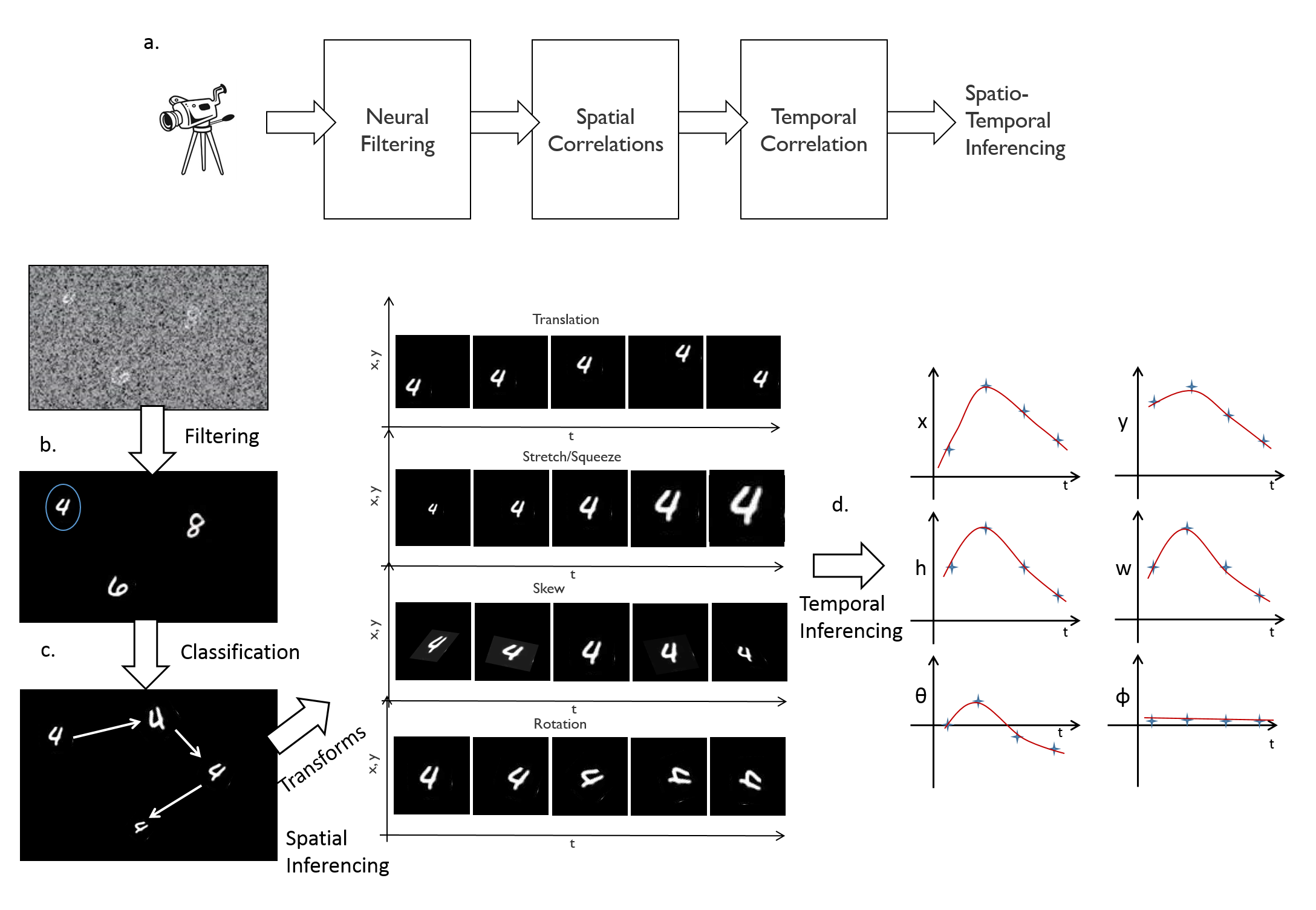}
 \end{center}
 \caption { \label{fig:1} 
a. Proposed neural filtering and processing architecture and the data flow. b. Neural filtering cleans up noisy images c. Spatial inferencing including classifications and transforms that identify features and extract its position, size, and degree of rotation and skew. d. Temporal inferencing uses the spatial correlations to generate equations of motion of the identified features through real-time learning. }
 \end{figure} 
	
\section{COGNITIVE IMAGE PROCESSING ARCHITECTURE: DESIGN AND PROTOCOL}

Cognitive image processing, being a broad term can mean different things in different contexts. As examples, for social media networks, the goals are primarily recognizing and mapping faces in an image to users, whereas in a self-driving automotive it would be to recognize street signs and lane markings in variable environmental conditions. Indeed all such ``cognitive tasks'' proceed from an application goal to most optimized neural network designs to achieve those tasks.

In our present work we define the goal of a smart camera platform to be able to identify and track a user defined feature from the camera's video feed. We want to formalize the scope of our design to able to perform the following tasks:

\begin{itemize}
	\item Extraction of clean signal from the noisy images taken by the camera, including various distortions such as haziness, speckles, fading etc.
	\item Identification of a pre-defined feature in an image frame, and extraction of spatial information of the feature, such as it's position, size, rotation etc. which can be considered as the canonical variables of motion for the feature.
	\item Extraction of the generative functions underlying the variables of motion and being able to predict the expected evolution of these variables.
\end{itemize}

Our goal is not so much as to identify/classify a large set of features in a collection of images, which is the most popular task in neural image processing literature, but rather user-defined feature tracking in real time operation of a camera.

The proposed architecture for this processing task is shown in fig.\ref{fig:1}a. We assemble three separate blocks which build the logic of processing listed above. This modularity of functionality allows us to select, design, and optimize each of these blocks independent of each other. Since this is an ongoing project, we have chosen to implement these three blocks using the following three types of neural networks, and the final design is subject to optimization as the platform matures. 

\begin{itemize}
	\item Neural Filtering: Reservoir Computing, particularly Echo-State Networks (ESNs).
	\item Spatial Inferencing: Enhanced Region-based Convolutional Neural Networks (R-CNNs).
	\item Temporal Inferencing: Hierarchical Temporal Memory (HTM).
\end{itemize}

The ESN based neural filters extract a signal from the noise by learning to invert the time-dependent distortions in the images (fig.\ref{fig:1}b) introduced by the sensor and it's operational limitations. Enhanced region-based CNNs extract the feature and its variables of motion from samples of the camera video feed, in form of a spatial correlation tuple. This tuple consists of the following quantities: $\{Tag, p, x, y, h, w, \theta, \phi_x,\phi_y\}$. The Tag item identifies the feature, in case we want to develop a multi-feature tracking camera, $p$ represents the probability of feature being in the image, $(x,y)$ represent the most probable position of the feature in the image frame, $(h,w)$ represent the size of the feature in the image frame, $\theta$ is the rotation of the features with respect to the initial position, and $(\phi_x,\phi_y)$ represents the skew in the feature due to rotation in the 3D and its projection on the 2D image frame. A collection of HTMs then develop equations of motions from a time-sequence of these tuples.

\section{SPATIO-TEMPORAL NEURAL NETWORKS}

We briefly describe the three types of neural networks we use in this work.

\subsection{Reservoir Computing}

Reservoir Computing (RC) is a model of computing built to handle time-varying data and learn correlations and patterns in them \cite{jaeger_technical_2001,maass_real-time_2002}. RC consists of a collection of loosely coupled neurons with structural recurrence (feedback-loops) on which a time-varying signal $u(t)$ is imposed. The state of the reservoir $x(t)$ as a result at any time is given by a combination of previous states $x(t-1)$ and the input signal $u(t)$. The new state $x(t)$ as a result is an integral of the signal over its previous samples $u(t),u(t-1),u(t-2)\ldots$ This gives rise to temporal-memory in the reservoir (also called echo-states, past signal being echoed throughout the network), and allows it to be used as a temporal correlator \cite{jaeger_harnessing_2004}. To prevent runaway positive feedback of signal (energy) driving the reservoir dynamics to chaos, a decay term is incorporated which fades the effect of the previous samples during the integration (dissipation due to friction). The dynamical equation for the reservoir shown in fig.\ref{fig:2}a is given by:

\begin{eqnarray}
\frac{dx}{dt} &=& -\eta x + \alpha\tanh(W^{self}x(t)+W^{in}u(t)+W^{fb}y(t)) \\
\label{eq:rc-state}
y(t) &=& W^{out}x(t)
\label{eq:rc-learn}
\end{eqnarray}

The output of the RC is extracted by a weighted sum of all the states of the reservoir. During learning, only the weight $W^{out}$ is adjusted using a simple regression. The training can be performed in both batch as well as online modes.

It should be noted that the RCs are limited by the choices of network parameters such as the size of the network, decay rate, spectral radius of $W^{self}$ etc. which fix the characteristic window size and the envelope shape of the temporal integration in the reservoir. Therefore, we choose HTMs rather than RCs for temporal inferencing, since we do not want to a priori fix these critical characteristics. We have chosen to use RC for the task of neural filtering, as it has been used to successfully perform an equivalent task in digital communication: channel equalization \cite{boccato_echo_2011}. This neural filtering can be applied at both pixel or feature level.

\subsection{Convolutional Neural Networks}

Convolutional Neural Networks (ConvNets or CNNs) are an application of learning techniques applied to images, and are an extension to traditional image processing techniques. The dataframe is assumed to have a meaningful 2D spatial relation, i.e. pixels close to each other are highly correlated and compose parts of an image \cite{krizhevsky_imagenet_2012}. 

The central operation of CNNs is the convolution operation (fig.\ref{fig:3}a), where a ``kernel'' $U_{j \times k}$ matrix is multiplied to an image $A_{m \times n}$, ($m>j, n>k$) in a sliding manner. Mathematically it is represented as the linear operation (in spatial domain) as:

\begin{equation}
B = U*A
\label{eq:cnn_spatial_domain}
\end{equation}

It should be noted that the result of the convolution operation $B$ is often trimmed, using a pooling operation, where at any point, a few adjacent pixels are represented using a single pixel data. As an example, a pooling window of $3\times3$ will reduce a $100\times100$ image to a $33\times33$ image. 

In traditional image processing \cite{jain_fundamentals_1989}, the $U$ is pre-calculated and designed to perform a specific operation (say edge detection, blurring, contrast enhancement etc.). In CNNs, these kernels are obtained through backpropagation techniques. The central purpose of most CNN applications are to classify features present in the image into target categories. As an example, consider an image consists of hand-written number and letters. The CNN can be trained to extract the number or a letter occurring in the image. 

CNNs in general need to be trained in a supervised setting, i.e. during its training it needs to be told the error in classification at its output, so that the CNN can adjust the convolutional kernels. Therefore, training of the targeted feature set for our platform will need to be performed offline. However, once trained on that particular feature set, the rest of the platform can learn, adapt, and predict online and in real-time.

\subsection{Hierarchical Temporal Memories}

HTMs are a neural network specialized for time-series data/temporal modeling, prediction and anomaly detection. Its design and learning methodology has been developed keeping in mind the design and operation of neocortices in sentient organisms, which perform the task of temporal integration of information and prediction seamlessly and in real time \cite{hawkins_intelligence:_2007,hawkins_theory_2017}.

In HTM, the neurons or ``cells'' are organized in a 3D stack as shown in fig.\ref{fig:4}a. The structural interconnectivity in these cells is designed to be highly recurrent (i.e. with feedback loops) which allows temporal correlations between samples to persist in the network hence the name temporal memory. In HTM a Hebbian like rule is implemented for learning, where the connections between the cells (dendritic connections) is formed or unformed depending upon the degree of correlations between two cells during the operation of the network \cite{hawkins_why_2016}. In fact, there is no separate learning phase for the complete network and the dendritic connections are updated online and in real-time during the processing \cite{cui_continuous_2016}, which makes it suitable for temporal inferencing.

Since the temporal memory size can be dynamically changed in the HTM, unlike RC, it is a more suitable network for the task of temporal correlation.

\section{NEURAL FILTERING AND PROCESSING TASKS}

\subsection{Filtering by Learning}

 \begin{figure} [h]
 \begin{center}
 \includegraphics[width=5in]{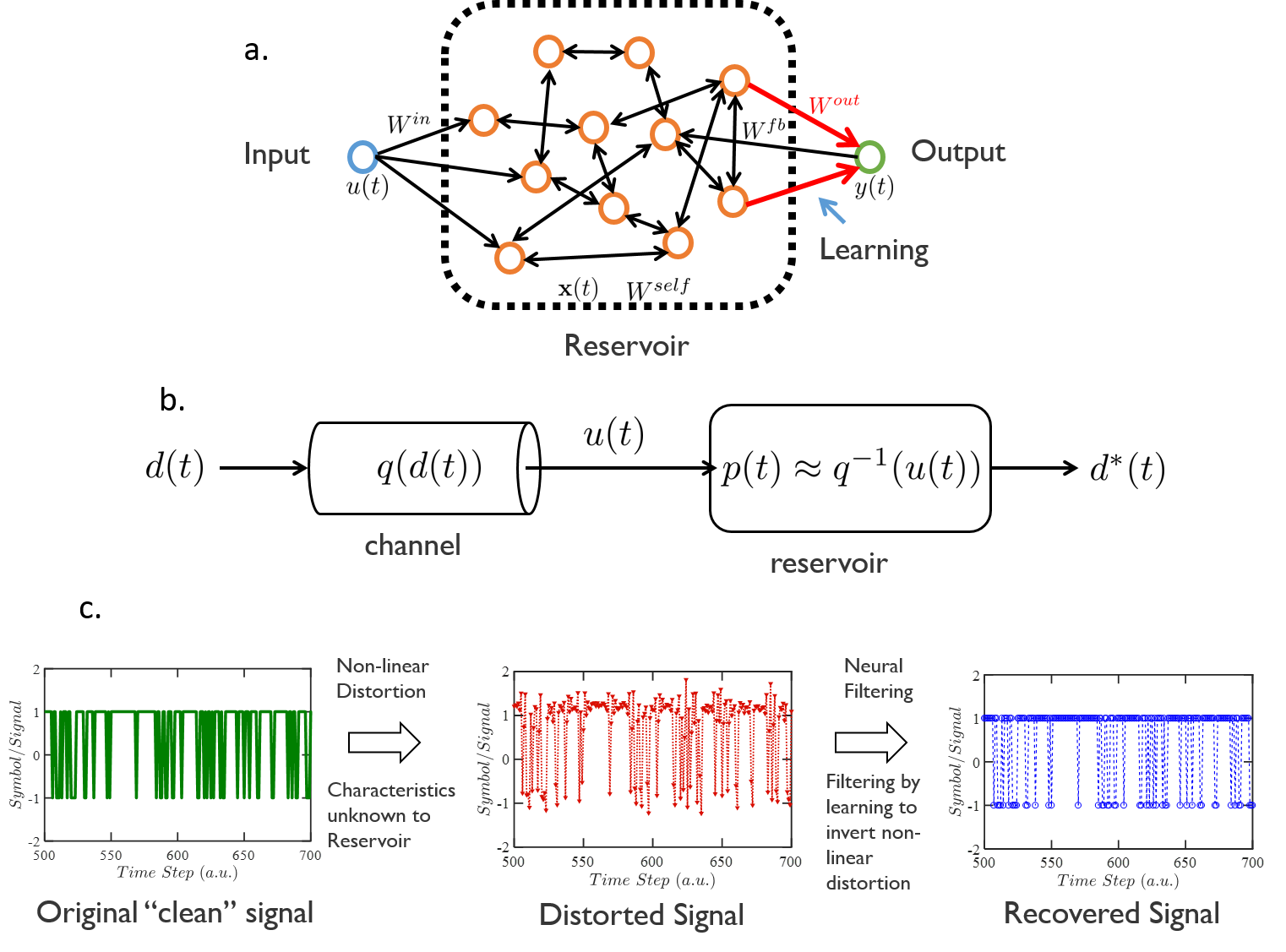}
 \end{center}
 \caption { \label{fig:2} 
a. Reservoir Computer schematic. A reservoir is composed of loosely and randomly coupled set of neurons whose weights are not changed during the training. The interconnectedness of the reservoir allows a given time sample of the signal to persist in time within the network and get integrated with newer samples. The only weight changed during the training is $W^{out}$ which can be adjusted to learn the correlation of the reservoir for specific tasks. b. Central idea behind neural filtering. The channel distorts the signal $d(t)$ by a non-linear function $q(d)$. The reservoir is trained to invert $q(d)$ to reconstruct the original signal. c. A simulation example showing the recovery of the signal in time.  }
 \end{figure} 

The neural filtering presently implemented is inspired by an equivalent task in digital communication called channel equalization. The source of the information produces a bitstream, which passes through a channel and is distorted in a non-linear fashion. Therefore it is not possible to simply apply a linear filter to extract the pure signal component. Instead, the effect of channel has to be reversed by creating an inverse of the channel characteristics, typically done in frequency domain using principal component analysis based methods.

We use the temporal modeling capabilities of RC to perform this task, where the channel is the detector which introduces distortions and noise due to its physical limitations. The process is illustrated in fig.\ref{fig:2}b, where the detector (channel) introduces distortion and noise on the original signal $d(t)$ through an unknown function $q(z)$ and as a result the output is given by:

\begin{equation}
u(t) = q(d(t))
\label{eq:rc-distorted}
\end{equation}

During the training process, the RC is fed this distorted signal $u(t)$ as its input and targeted to reproduced the original $d(t)$ by minimizing the difference from its actual output $d^*(t)$. This in effect turns the reservoir into a inverse modeler:

\begin{eqnarray}
p(z) &=& q^{-1}(z) \\
d^*(t) &=& p(u(t))
\label{eq:rc-recovered}
\end{eqnarray}

Once the inverse model has been learned, the reservoir can now be used for neural filtering tasks as shown in fig.\ref{fig:2}c, where an original clean signal can be recovered with high accuracy (normalized root mean square error (NRMSE) $<10^{-3}$).

\subsection{Spatial Inferencing}

Spatial inferencing involves extracting the probability of a feature being present in the image along with its variables of motions. However, regular CNNs do not provide any of the variables of motion for the identified feature in the image. For the purpose of tracking an object in an image frame, it is important that the CNN provides the positional and transformation information, i.e. the full tuple described previously, for the temporal inferencing. 

\begin{figure} [h]
 \begin{center}
 \includegraphics[width=5in]{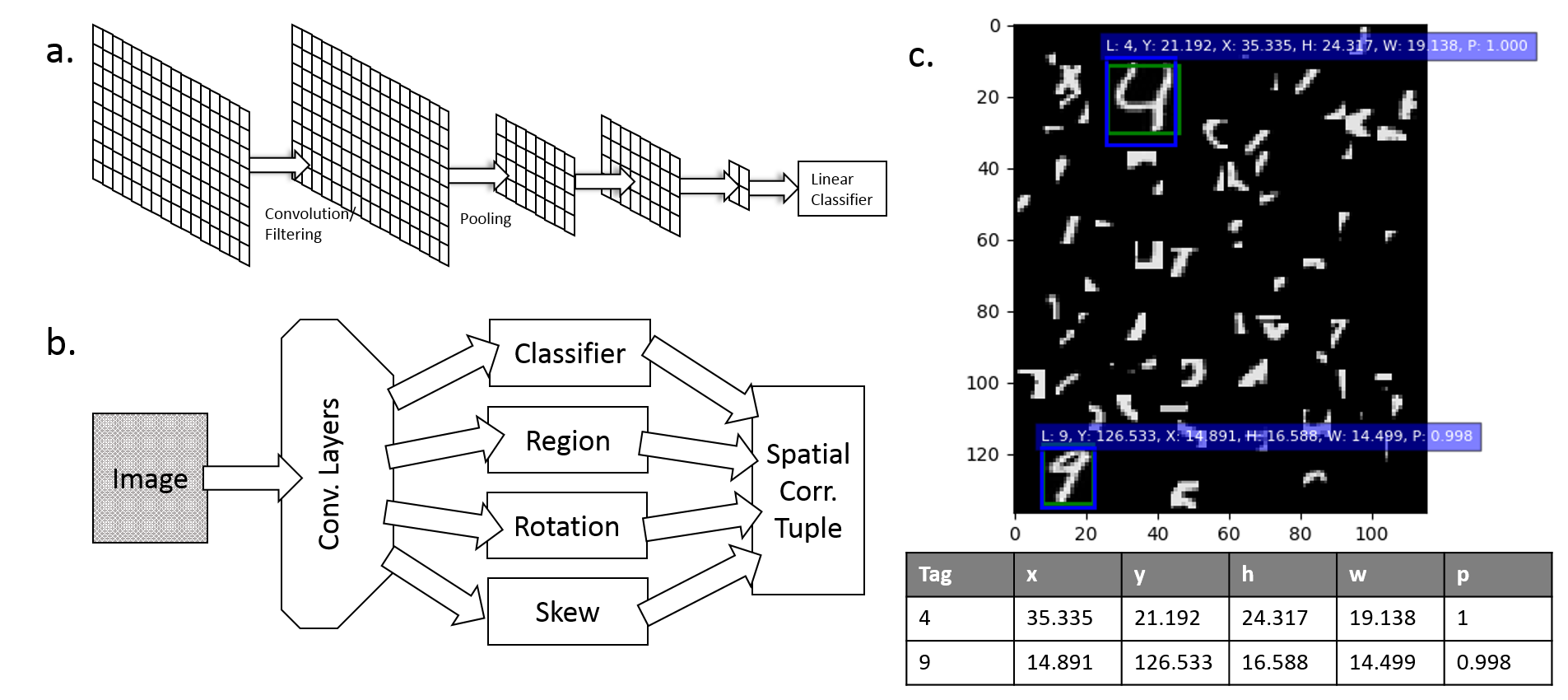}
 \end{center}
 \caption { \label{fig:3} 
a. Convolutional Neural Networks showing convolution and pooling operations. Feature maps are extracted at each convolutional layer and compactified using pooling. b. An enhanced CNN using extra region, rotation and skew flows that together build the spatial correlation tuples. c. An example of a region based extraction of feature location, size, and probability (see \cite{liang_tf-faster-rcnn:_2018} for an implementation). The implemented R-CNN can extract multiple features and their position and sizes with high accuracy, as can be seen from the table. Rotation and skew flows can be implemented in a similar modular fashion.}
 \end{figure}

To achieve this, we apply a region based CNN (R-CNN) which also extracts the most probable smallest bounding box that contains this feature within the image frame \cite{gidaris_object_2015}. This can be achieved by using sliding windows of various sizes exhaustively over the image frame or through a combination of segmenting the image into various region proposals/regions of interest, and then searching within the regions \cite{ren_faster_2015,he_mask_2017}. In addition, rotation and skew  information can also be extracted by geometrically transforming the regions of interests from straight to rotated rectangles with best search incidences to infer the angle of rotation \cite{liu_rotated_2017}, and using parallelograms instead of rectangles to obtain the skew angles (fig.\ref{fig:3}c). 

Therefore, the enhanced R-CNN allows us to form the spatial correlation tuples $\mathrm{\{Tag,p,x,y,h,w,\theta,\phi_x,\phi_y\}^T}$. Storing the tuple information from the previous image sample in a video allows us to obtain the next sample's tuple faster compared to a purely static image frames because in an time sequenced image samples, physical objects shift only by a finitely small amount. In addition, if the camera is embedded on a mobile platform, sensor data from the platform's motion can be used to compensate for the drift by developing more region proposals.

\subsection{Temporal Inferencing}

\begin{figure} [h]
 \begin{center}
 \includegraphics[width=5in]{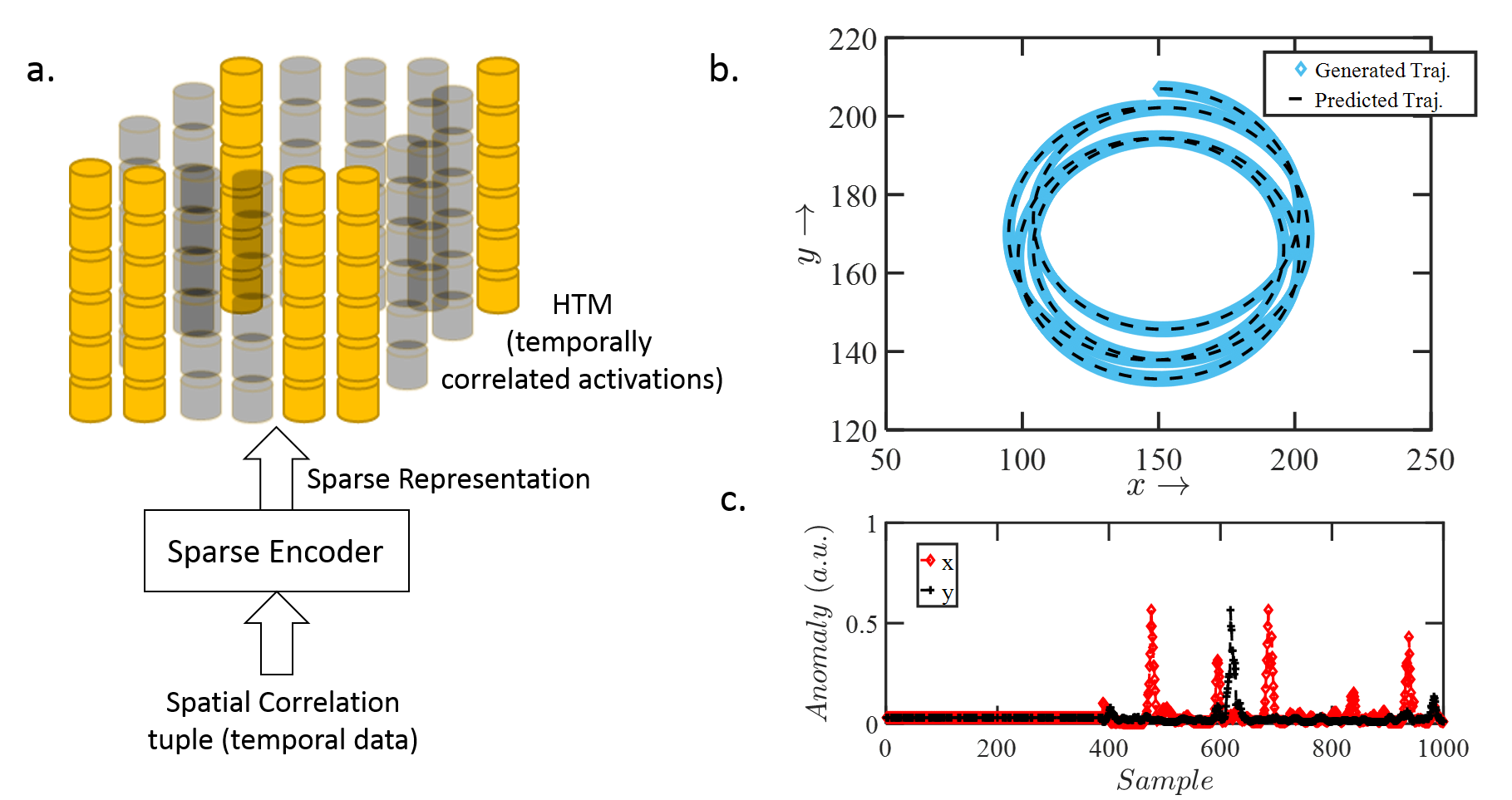}
 \end{center}
 \caption { \label{fig:4} 
a. HTM with temporally correlated activations. b. An HTM based prediction of time series data. A complex trajectory of motion generated by an underlying model not known to HTM can be learned and predicted simultaneously from  the time ordered samples of the trajectory data stream. c. Anomaly between predicted model and actual data.}
 \end{figure}

Temporal inferencing involves developing a generative model for the time-sequenced samples of the spatial correlation tuples. The seven variables of motion are separately fed to seven separate HTMs (fig.\ref{fig:4}a) through a sparse encoding layer and the cortical learning algorithm is applied to the individual data streams. In HTMs, the learning and prediction are interleaved together, which makes it an online and real-time process. 

Fig.\ref{fig:4}b shows an example of temporal learning and prediction relevant to our application. A trajectory in $x-y$ plane is generated using a deterministic set of functions. Individual samples of the coordinates $\{x(t),y(t)\}$ are provided as a list to two separate HTMs. It can be seen that the HTMs can learn and model the trajectory successfully, which proves the validity of the approach. Fig.\ref{fig:4}c shows an anomaly map for the samples, where anomaly is defined as the difference between the actual trajectory and predicted trajectory. There is an initial period of 0 anomaly in the samples, as the HTMs learn and build the model, but after that the HTM runs in a simultaneous predictive-generative mode. For very long periods of operation, the predictive capability of the HTMs improve for smooth trajectories and any ``sudden change'' in the trajectory will result in large amount of anomalies, which can be reported as an ``alarming event'' to the user of the platform.

This approach to model the positional ($\{x(t),y(t)\}$) variables of motion can also be extended to the extensional (size $\{h(t),w(t)\}$), rotational ($\theta(t)$), and skew ($\{\phi_x(t),\phi_y(t)\}$) variables of motion as well. Generative models of the full tuple can then be used by a cyber-physical system (CPS) in the lines of ``sensory neuron $\rightarrow$ motor neuron'' connection in sentient organisms for applications such object tracking, locomotion etc.

\section{HARDWARE FOR NEURAL NETWORKS}

Most state-of-the-art and advertised neural networks are developed and run on Graphical Processing Units (GPUs), primarily due to their ability to perform linear algebra operations such as dot products directly in hardware. This illustrates the mismatch between the von-Neumann architecture with its separation of memory and logic and the needs of neural network designs. In Neural networks, the memory (synaptic weights) is spread throughout the computational fabric, and the basic unit of operations, dot-product and thresholding do not map directly to NAND gate based logic designs. 

As a result there is a concerted effort to re-imagine hardware designs that can closely match the neural architectures \cite{misra_artificial_2010}. Large organizations such as Google, Intel, IBM, Micron etc. are already building prototype Deep Learning accelerators that implement hardware architectures matching neural network designs at close-to-code level. This effort is still broadly at the level of emulation of neural networks using Boolean computing units (CMOS transistors) with better logic primitives (hardware dot-product etc.). There is also a parallel effort in developing custom designed CMOS based ``digital neurons'' implemented in FPGAs as well as exploring nano-materials such as spintronics and various memristive materials, and physical phenomena governing them to develop better ultra-compact hardware units that encompass some/all physical characteristics necessary to implement a neuron and synaptic weight adjustment during learning directly. 

It is largely expected that the uptake of hardware neurons will happen first in the emulation modes, where CMOS based ``digital neurons'' will implement large neural networks. Eventually these neural fabrics may be enhanced and scaled using the nano-materials based neurons, sparking a new Moore's Law of electronics.

\begin{figure} [h]
 \begin{center}
 \includegraphics[width=5in]{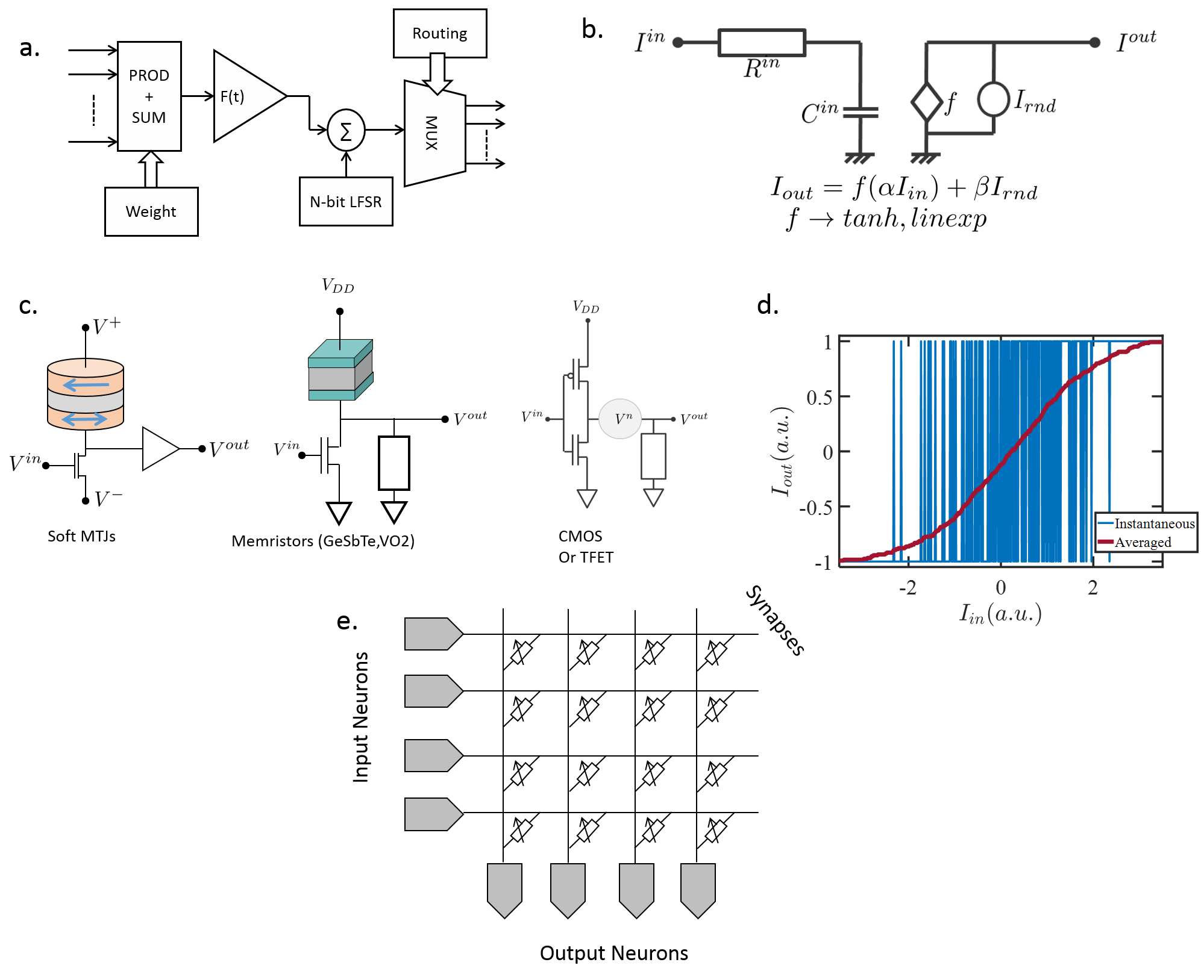}
 \end{center}
 \caption { \label{fig:5} 
a. An example digital hardware neural model that can be implemented in FPGAs or ASICs. b. General hardware neural model. c. The neural model can also be obtained directly by ultra-compact hardware units built using emerging nano-materials that show stochastic switching: soft-magnets, memristors, and subthreshold TFETs. d. An example switching characteristics of the neural model implemented using the soft-magnet MRAM, which encodes the neural output in the average response (red averaged current) which has a logistic like activation. e. Representational layout of hardware neural networks using a memristive cross-bar array as synaptic connections, while the neurons sit at the periphery. The individual interconnections (synaptic weights) can be programmed during training phase by programming the controllable resistance of the memristor connecting the row and the column between the input and output neurons. Deep networks can be built by cascading/stacking this design repeatedly.}
 \end{figure} 

\subsection{Field Programmable Gate Arrays (FPGAs) and Digital Neurons}

FPGAs are the platform of choice for prototyping new architecture designs. They are composed of large arrays of programmable Boolean gates and interconnection networks which can be used to build basic building blocks from bottom up such as ADDER, MUX, register, LFSR, Encoder and Decoders, Content Addressable Memories etc. It is possible to build complex logic designs in FPGAs by combining these blocks programmatically using well developed workflow based on Hardware Description Languages (HDLs) (Verilog, SystemC, SystemVerilog) which describe the dataflow through the logic blocks (also called register-transfer level design). 

A possible example of a digital neuron and synapse that can be designed and implemented on FPGAs is shown in \ref{fig:5}a. The synaptic weights are stored in a memory buffer and used to create the argument $(z=wx +b)$ for neural activation using the PROD+SUM block, which can be implemented using separate MULTs (such as Wallace and Dadda trees) and ADDERs (Ripple Carry, Koggstone etc.) or using Fused Multiply-Add (FMA) cores. The thresholding function $f(wx +b)$ can be implemented either using low order power law approximations of $f(z)$ function, or using a lookup table based method, in which the numerical value of the argument $z$ is mapped to the memory address where the result of the computation is stored (also called hashing). Linear feedback shift registers (LFSRs) can be used to generate random numbers if stochasticity is necessary for the computation, and the fan-out routing logic between the various neurons is implemented using a MUX.

From the above design it is clear that the performance of these digital neurons can be improved significantly if the synaptic weighing and addition, thresholding, and routing can be implemented using compact hardware units whose physics matches these requirements, since implementing these blocks in FPGAs or ASICs using CMOS based Boolean gates still require a lot of silicon area and associated energy costs. Emerging nano-materials based technology such as spintronics \cite{debashis_experimental_2016,camsari_implementing_2017} and memristors \cite{misra_artificial_2010,du_reservoir_2017, kulkarni_memristor-based_2012} are of particular interest for such applications.

\subsection{Emerging Technologies: Spintronics and Memristors}

Spintronics and memristors are emerging technologies that do not use electronic charge as the state variable of computation, but rather use direction of magnetization of a thin magnetic layer (expressed as a vector $\hat m$), and internal chemical configuration (expressed by chemical potential $\mu$) respectively. These variables effect the electrical readout of a device built using them, typically by controlling the resistance of the structure. These state variables can be manipulated by passing a current through them enabling a write mechanism. Most prominent use case for these materials at present is in non-volatile memory, where spintronics in particular has made a niche for itself as a possible successor of NAND flash. We describe how these materials can be used to build hardware units for neurons. We will focus on spintronics, but the central principles can be transferred to memristive systems as well.

A general hardware neuron model is shown in fig.\ref{fig:5}b. The input side has a natural delay line built from a resistor and a capacitor, which stores the state of the cell. The charge on the capacitor $Q_{in}$ determines whether the device will switch ($Q_{in}\ge Q_c$) or not ($Q_{in}<Q_c$). The output transfer function $f(\alpha\frac{dQ_{in}}{dt})$ is a non-linear one, typically $\tanh(z)$ but can be others like  rectified linear units (ReLU) etc. There is also a additive white Gaussian noise (AWGN) current source in parallel to the transfer function and generates the output noise for the neuron. This noise is particularly critical in implementing the stochasticity that is needed in modern neural learning and optimization algorithms. The characteristic switching function is given by:

\begin{equation}
I_{out} = f(\alpha \frac{dQ_{in}}{dt}) + \beta I_{rnd}
\label{eq:general-neuron}
\end{equation}

This generalized behavior is demonstrated by a variety of materials (fig.\ref{fig:5}c). Magnetic Tunnel Junction (see \cite{kent_new_2015} for a comprehensive review) built using soft-magnets or super-paramagnets are a particularly attractive option for implementing stochastic neurons, even though deterministic neurons using hard-magnets have also been proposed, e.g. see \cite{quang_diep_spin_2014}. The degree of hardness of magnets is characterized by the height of its internal energy barrier separating the two minimum energy states. This energy barrier is a function of it material parameters (saturation magnetization $M_s$, uniaxial anisotropy field strength ($H_k$), and volume $\Omega$) and is given by:

\begin{equation}
U = \frac{M_sH_k\Omega}{2}
\label{eq:magnet-stability}
\end{equation}

This determines the state retention time (degree of hardness) that can be captured by:

\begin{equation}
\tau = \tau_0\exp(\frac{U}{kT})
\label{eq:magnet-time}
\end{equation}

where attempt frequency $\tau_0$ is typically $0.1-1\ ns$. Therefore for $U\le 1kT$, the retention time of the magnet is a few $ns$ or less. An input current $I_{in}$, converted to a spin transfer torque by passing through a polarizing layer, if strong enough can align the magnet to a preferable direction. Therefore, if we sweep the $I_{in}$, the instantaneous response of the device is given by:
\begin{equation}
I_{out} \propto m^{out}_{z} = \rm{sgn}(\tanh(\kappa m^{in}_{z}(I_{in}) + \rm{rnd}(-1,+1))
\label{eq:spin-device}
\end{equation}

However, the long time average of the of the output $I_{out}$ follows a $\tanh(z)$ characteristic, and therefore matches the requirement of a binary stochastic neuron, which is the general neuron model used in most deep learning algorithms. The metallic leads of the device can automatically add the currents, whereas the synaptic weights can be implemented by controlled conductors, since $I = GV$. A hybrid crossbar array of programmable memristors can compactly implement these programmable synaptic connections at the junctions and the neurons at the periphery (fig.\ref{fig:5}e).

These networks can then be used in a variety of neural network designs, including combinatorial optimization solvers, Boltzmann Machines (BMs) and Deep Belief Networks (DBNs), and Reservoir Computers (RCs), some of the central types of neural networks we have discussed here. Please see the refs. \cite{ganguly_evaluating_2016,camsari_stochastic_2017,sutton_intrinsic_2017,ganguly_reservoir_2017} for more detailed exposition and applications.

\section{SUMMARY} 

In this work, we presented an architecture for a smart camera platform that will eventually employ a combination of best-in-class neural networks to perform its principle tasks: a) neural filtering, b) spatio-temporal inferencing. The platform has been designed with an application of user defined feature tracking and prediction in mind. The platform is generic enough to be applied to a wide variety of established and emerging fields including defense, automated vehicles, and autonomous sensor networks.

We described the three classes of neural networks used in the work and demonstrated learning and prediction tasks individually from these networks. Using Reservoir Computing, we demonstrated signal recovery by inverse modeling of a noisy detector (fig.\ref{fig:2}). Using enhanced CNNs we demonstrated extraction of a feature, along with its variables of motion within an image frame (fig.\ref{fig:3}). Using HTM, we demonstrated learning and prediction of the trajectory of a feature in the image frame (fig.\ref{fig:4}).

We further pointed out possible hardware implementation of neurons for energy-efficient processing that can be embedded close to the image sensor. We discussed how digital neurons can be implemented on present day FPGAs and how it is possible to leverage the physics of emerging nano-materials to enhance conventional CMOS based computing in directly implementing neural networks in large scale circuits. 

\section*{ACKNOWLEDGMENTS}
This work was supported in part by the NSF I/UCRC on Multi-functional Integrated System Technology (MIST) Center IIP-1439644, IIP-1738752 and IIP-1439680. We would like to thank Dr. Kerem Yunus Camsari for useful discussions on soft-magnet based stochastic neurons.

\bibliography{SpatioTemporal} 
\bibliographystyle{spiebib} 

\end{document}